%% file: main.tex
\documentclass[10pt,twocolumn,letterpaper]{article}

\usepackage{cvpr}              

\input{preamble}

\definecolor{cvprblue}{rgb}{0.21,0.49,0.74}
\usepackage[pagebackref,breaklinks,colorlinks,allcolors=cvprblue]{hyperref}

\usepackage{adjustbox}
\usepackage{multirow}
\usepackage{amsmath}
\usepackage{amssymb}
\usepackage{booktabs}
\usepackage{colortbl}
\definecolor{mintbg}{rgb}{.63,.79,.95}
\colorlet{lightmintbg}{mintbg!20}

\def\paperID{18093} 
\def\confName{CVPR}
\def\confYear{2025}

\title{Precise Event Spotting in Sports Videos: Solving Long-Range Dependency and Class Imbalance}

\author{Sanchayan Santra$^{1}$ \quad Vishal Chudasama$^{1}$ \quad Pankaj Wasnik$^{1}$ \quad Vineeth N Balasubramanian$^{2}$ \\
$^{1}$Sony Research India \quad  $^{2}$Indian Institute of Technology Hyderabad\\
{\tt\small \{sanchayan.santra, vishal.chudasama1, pankaj.wasnik\}@sony.com, vineethnb@cse.iith.ac.in}}

\begin{document}
\maketitle
\input{sec/0_abstract}
\input{sec/1_intro}

\section{Related Works}
\label{sec:related}
The concept of sports event spotting first appeared in SoccerNet \cite{soccernet} with the aim of identifying specific events or actions at exact locations. The work by \cite{e2espot} expanded the event spotting task to precise event spotting (PES), introducing significant differences in the necessary precision of predictions. PES is constrained to the tolerance of few frames only. 
Recently, several PES methods \cite{e2espot, baidu, yahoo-icip, astra, comedian, T-DEED, uglf} have been proposed that focus on various sports videos. E2E-Spot~\cite{e2espot} employed a CNN-based backbone with a Gated Shift Module (GSM)~\cite{GSM} for temporal modeling and a GRU~\cite{gru} for producing per-frame prediction. Even though it is effective on the Tennis and Figure Skating datasets, it struggles on the more challenging SoccerNet V2 dataset~\cite{SoccerNetv2} due to its simplistic temporal information extraction scheme using GSM. 

Recently, T-DEED~\cite{T-DEED} proposed improvement on top of E2E-Spot~\cite{e2espot} to address issues arising from tight tolerances in fast-paced sporting events and achieved SOTA performance on the Figure Skating and Figure Diving datasets. The method of \cite{baidu} revamped the feature extraction using a group of five networks pre-trained on Kinetics-400 and fine-tuned on SoccerNet V2. Several other methods use the features extracted by this method instead of raw video frames. For instance, Soares et al.~\cite{yahoo-icip} utilized these features and developed a U-Net- and transformer-based trunk for making predictions. ASTRA~\cite{astra} utilizes the same features but also incorporates audio features for complementary information. In contrast, COMEDIAN~\cite{comedian} operates in 3 stages to train their network: 1) training only the spatial part; 2) training the spatial and temporal parts using knowledge distillation; 3) final tuning to optimize the classification scores. Recently, UGLF~\cite{uglf} proposed a unifying global local (UGL) module to disentangle the scene content into global features and locally relevant scene entities features in conjunction with a Vision-Language model.

\begin{figure*}[t!]
    \centering
    \includegraphics[width=\linewidth]{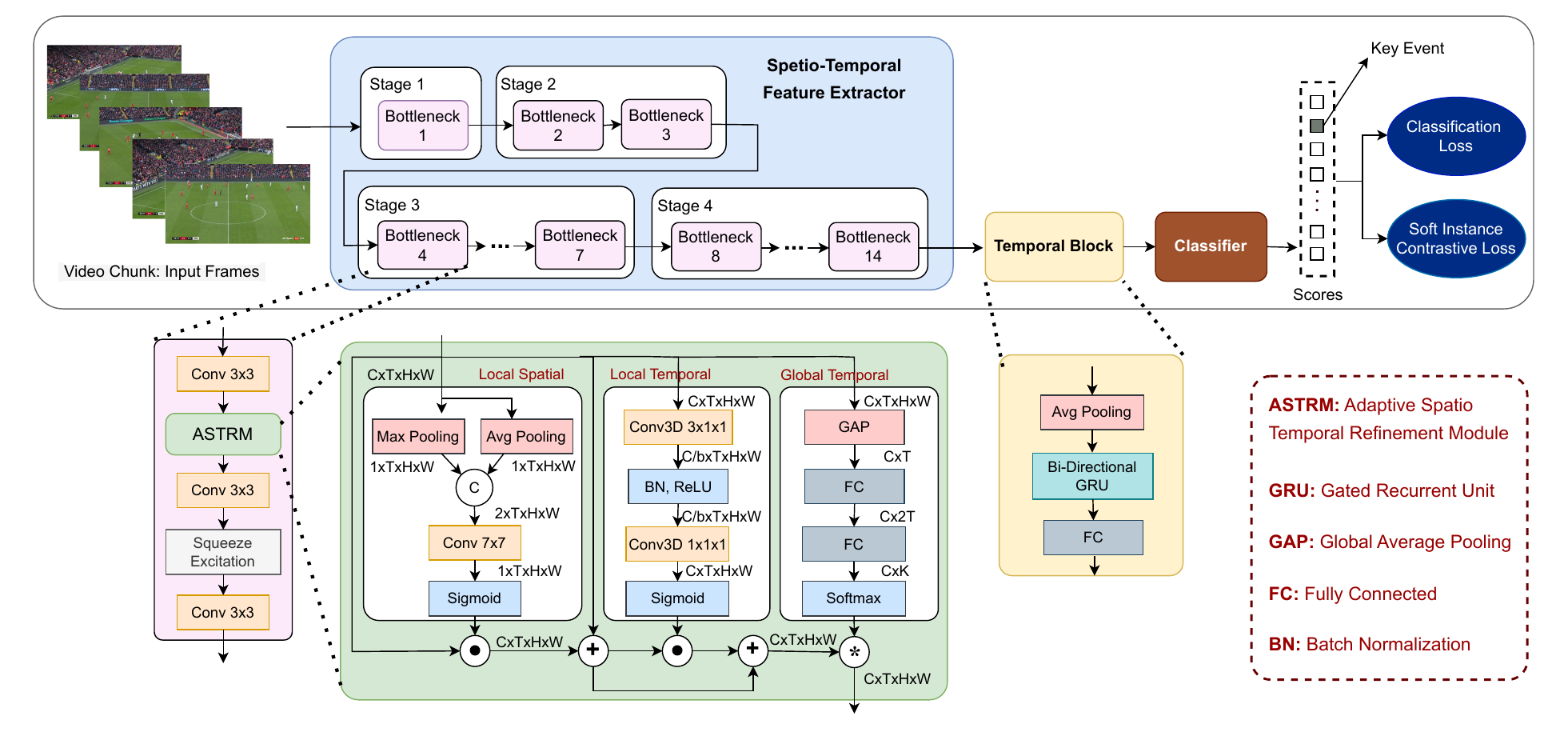}
    \caption{\textbf{Network design of the proposed framework.} The framework composed of a spatio-temporal feature extractor and a temporal block for capturing long-range dependency before the classifier. In each bottleneck block we add ASTRM after the first conv. ASTRM further enhances the features with local spatial, local temporal and global temporal information. The network is trained with SoftIC loss in addition to the classification loss to handle class imbalance.}
    \label{fig:framework}
\end{figure*}
The above-mentioned methods had mainly addressed the long-range temporal dependency issue by utilizing pre-trained features and transformers. They have not adapted their feature extractors to capture both spatial and temporal information effectively. Additionally, these methods did not specifically deal with the class imbalance, which led to poor performance on underrepresented classes. In our framework, we proposed ASTRM, which helps the feature extractor to augment the features with relevant spatial and temporal information. Furthermore, we introduce SoftIC loss to address the class imbalance issue and employ the ASAM to properly train the proposed framework. With these advancements, the proposed framework outperforms existing event-spotting methods on various sports datasets, particularly in settings with tighter tolerances.

\section{Proposed Methodology}
This section presents our methodology in detail, starting with the problem statement and then presenting the subsequent details as follows.
\subsection{Problem Statement}
The task of event spotting is defined as follows: Given a video with $N$ frames $\mathbf{x}_1, \mathbf{x}_2, \ldots, \mathbf{x}_N$ and a set of possible events $\{c_1, \ldots, c_K\}$, the goal is to predict frames where an event occurs along with it's class: $(t, \hat{y}_t) \in \mathbb{N}\times \{c_1, \ldots, c_K\}$. A prediction is deemed correct if it falls within $\delta$ frames of the actual occurrence and accurately identifies the event class. Since it is not possible to process the entire video at once, the processing is done in chunks of a predefined size. Each chunk, represented as $\mathbf{X} \in \mathbb{R}^{C\times T_s\times H\times W}$, consists of $T_s$ frames. We compute predictions for each frame, represented as $\mathbf{Y}\in \mathbb{R}^{T_s\times K}$, where $K$ is the number of classes. As the video is processed in segments, the predictions from each chunk of $T_s$ frames are aggregated and refined with a non-maximal suppression to get the final sparse prediction.
 
\subsection{Methodology}
\label{sec:method}
Figure~\ref{fig:framework} shows the network design of our proposed framework, taking a set of input frames and classifying whether each frame contains an event frame or not. The framework comprises three main parts: the spatio-temporal feature extractor, the temporal block, and the classifier. In the spatio-temporal feature extractor, we have selected the RegNetY~\cite{regnet} as the feature extractor due to its efficiency, surpassing commonly used networks like ResNet. This network is further enhanced with our proposed Adaptive Sptio-Temporal Refinement Module (ASTRM). This module is integrated into each Bottleneck block, refines the features by incorporating both spatial and temporal information. Both spatial and temporal information is necessary to properly identify events as validated by the experiments.
The refined features are then passed to the temporal block, which enriches them with long-term temporal information. The long-term information is necessary because in many situations, information from temporally distant frames is essential to accurately classify a frame. The spatio-temporal feature extractor alone can not accomplish this task. Finally, the feature maps are forwarded to the classifier, a simple linear layer that outputs the classification scores. Furthermore, we introduce the Soft Instance Contrastive (SoftIC) loss to effectively manage the imbalance among the class samples. This loss function operates by compacting the features and enhancing their separability. The important modules of the proposed framework are detailed in subsequent subsections.

\subsubsection{Adaptive Spatio-Temporal Refinement Module (ASTRM):}
The module enhances the extracted features by incorporating both spatial and temporal information. It composed of three blocks: Local Spatial, Local Temporal, and Global Temporal. We include the Local Spatial block because local spatial features also offer important cues about temporal information. Relying solely on temporal information can overlook these valuable insights. The spatial component is inspired by \cite{CBAM}, while the temporal aspect is adapted from \cite{TAM}. The refinement is done as follows:  
\begin{equation}
\Psi(\mathbf{x}) = ((\mathbf{x} \odot (1 + \mathcal{F}_s(\mathbf{x}))) \odot (1 + \mathcal{F}_t(\mathbf{x}))) * \mathcal{G}_t(\mathbf{x})
\end{equation}
where $\mathbf{x}$ is the input feature to be enhanced. $\mathcal{F}_s$, $\mathcal{F}_t$, and $\mathcal{G}_t$ denote the operation corresponding to the local spatial, local temporal, and global temporal block, respectively. $\odot$ denotes Hadamard Product, while $*$ denotes convolution operation. To elaborate, the input feature is weighted with information from the local spatial block and a residual connection, resulting in features enhanced with local spatial information. These enhanced features are further refined with local temporal information. Finally, spatially and temporally enhanced local features are then convolved with the kernel obtained from the global temporal block to make local enhancements using ``global" information. This dynamic kernel approach simplifies the network while maintaining effectiveness. Please refer to Fig.~\ref{fig:framework} for a visual representation.

In the Local Spatial block, the input features are pooled using Max (i.e., $f_{MaxPool}$) and Average (i.e., $f_{AvgPool}$) pooling operations along the channel dimension. The pooled features are then concatenated and convolved with a $7\times 7$ filter (i.e., $f_{conv}^{7\times 7}$). The pooling operation on the channel axis helps to emphasize informative regions and the subsequent convolution operation filters out non-informative portions from the features. Finally, the output is passed through a Sigmoid function (i.e., $\sigma(\cdot)$) to obtain the ``spatial'' feature weights (i.e., $\mathcal{F}_s(\mathbf{x})$) as follow:
\begin{equation}
\mathcal{F}_s(\mathbf{x}) = \sigma\big(f_{conv}^{7\times 7}\big(\big[f_{AvgPool}(\mathbf{x});f_{MaxPool}(\mathbf{x})\big]\big)\big).
\end{equation}

In the Local Temporal block, we use 3D convolution operations, namely $f_{conv}^{1\times 1 \times 1}$ and $f_{conv}^{3\times 1 \times 1}$, followed by the Sigmoid function, $\sigma(\cdot)$, with a focus on the temporal dimension. Spatially pooled features are not used in this block because they do not account for the different temporal changes at various spatial locations. The locally temporal enhanced features are defined as below:
\begin{equation}
\mathcal{F}_t(\mathbf{x}) = \sigma\big(f_{conv}^{1\times 1\times 1}\big(\text{BN}\big(\text{ReLU}\big(f_{conv}^{3\times 1\times 1}\big(\mathbf{x}\big)\big)\big)\big)\big).
\end{equation}

In the Global Temporal block, we adaptively create convolution kernels to improve the feature maps by considering temporal information across frames from all input frames. First, the features are pooled in the spatial dimension using a global average pooling operation (i.e., $f_{GAP}$), and then fully connected layers (i.e., $f_{FC}$) are used to compute the kernel with a global view of the data. Mathematically the kernel is
formulated as below:
\begin{equation}
\mathcal{G}_t(\mathbf{x}) = \sigma\big(f_{FC}\big(f_{FC}\big(f_{GAP}\big(\mathbf{x}\big)\big)\big)\big).
\end{equation}

\subsubsection{Temporal Block:}
The spatio-temporal feature extraction stage computes features containing local spatial, local temporal, and location-independent temporal information. Since we are interested in finding the temporal location of the events, we aggregate the features in the spatial dimension with average pooling before extracting long-range temporal information. We use temporal block because the feature extractor alone is not capable enough to capture the required long-range dependency, although it may help to some extent. As motivated in the introduction section, we need to capture long-range temporal information to accurately predict events with precision. For this task, we selected Bidirectional GRU (Bi-GRU) as the temporal block because it effectively models sequences without overfitting the input data. The Bi-GRU is capable of quickly adapting to the data while maintaining its generalizability. Further experiments confirm this, demonstrating that the Bi-GRU performs well compared to more complex temporal blocks.

\subsubsection{Soft Instance Contrastive (SoftIC) loss):}
To address the class imbalance issue, we introduce a novel SoftIC loss which aims to make the features more compact and increase their separability.
It is inspired by the Instance Contrastive Loss (IC Loss)~\cite{opendet}, and is designed to work with \emph{mixup} augmentation~\cite{mixup}. 
The goal of IC Loss is to enhance the separation among the learned features, thereby improving classification scores when dealing with imbalanced data. For a batch of $N$ features computed from the input, the IC Loss is calculated as described in~\cite{opendet}.
\begin{equation}
\mathcal{L}_{IC} = \frac{1}{N}\sum_{i=1}^N \mathcal{L}(\mathbf{z}_i),
\end{equation}
\begin{equation}
    \mathcal{L}(\mathbf{z}_i) = \frac{1}{|M(\mathbf{c}_i)|} \sum_{\mathbf{z}_j \in M(\mathbf{c}_i)} \log \frac{\exp(\mathbf{z}_i \cdot \mathbf{z}_j/\tau)}{\sum_\mathbf{z_k \in A(\mathbf{c}_i)} \exp(\mathbf{z}_i \cdot \mathbf{z}_k/ \tau)},
\end{equation}
where $\mathbf{c}_i$ is the label of $i$-th sample, $\tau$ is a temperature hyper-parameter, $M(\mathbf{c}_i)$ denotes the memory bank of class $\mathbf{c}_i$ and $A(\mathbf{c}_i) = M\backslash M(\mathbf{c}_i)$.

During training, we use \emph{mixup} augmentation~\cite{mixup} to regularize the model and improve its generalization. In this, two samples (and their labels) are combined in a convex manner to create a new sample (and its label) as follows.
\begin{eqnarray}
\tilde{x} = \lambda x_i + (1 - \lambda)x_j, \\
\tilde{y} = \lambda y_i + (1 - \lambda)y_j,
\end{eqnarray}
where $x_i$ and $x_j$ are two samples from the training data, and $y_i$ and $y_j$ are are their corresponding one-hot labels. The parameter $\lambda$ takes value in the rage $[0, 1]$. Consequently, the label ($\tilde{y}$) will not be in one-hot format; instead, it will reflect weighted contributions from each class.

ICLoss assumes that each sample has a single label. However, this assumption does not hold true when using \emph{mixup} augmentation. 
To account for the weight associated with each class in the label, we define \emph{Soft IC Loss} as follows:
\begin{equation}
\mathcal{L}_{SIC} = \frac{\sum_{(\mathbf{z}_i, w_{ij}), w_{ij} \neq 0} \mathcal{L}_{SIC}(\mathbf{z}_i, w_{ij})}{\sum_{(\mathbf{z}_i, w_{ij}), w_{ij} \neq 0} 1},
\end{equation}
\begin{multline}
    \mathcal{L}_{SIC}(\mathbf{z}_i, \omega) = \frac{1}{\omega|M(\mathbf{c}_i)|} \times \\
    \sum_{(\mathbf{z}_j, w_j) \in M(\mathbf{c}_i)} \log \frac{\exp(\mathbf{z}_i \cdot w_{j}\mathbf{z}_j/\tau)}{\sum_{(\mathbf{z_k}, w_k) \in A(\mathbf{c}_i)} \exp(\mathbf{z}_i \cdot w_{k}\mathbf{z}_k/ \tau)}. \label{eq:soft_ic_p2}
\end{multline}
where, $w_{ij}$ is the weight associated with the label of class $c_i$ for sample $\mathbf{z}_i$. Additionally in the memory bank $M(\mathbf{c}_i)$ we store the corresponding class weights along with the features.

\subsection{Training Strategy}
The proposed framework is trained end-to-end way using a combination of two loss functions: Binary Cross Entropy loss ($\mathcal{L}_{BCE}$) as the standard classification loss and the proposed SoftIC loss ($\mathcal{L}_{SIC}$). The overall loss function ($\mathcal{L}_{final}$) can be expressed as below:
\begin{equation}
    \mathcal{L}_{final}(\cdot) = \mathcal{L}_{BCE} + \lambda_{SIC}\cdot \mathcal{L}_{SIC}.
\end{equation}
Here, $\lambda_{SIC}$ denote the weighting constant that adjust the contributions of different loss functions. With the total loss, proposed network is trained using the AdamW~\cite{adamw} optimizer with, Adaptive Sharpness-Aware Minimization (ASAM)~\cite{asam}. 

Utilizing advanced optimization techniques can significantly impact the performance of the trained model, yet not much focus is given to this aspect. In \cite{yahoo-challenge}, the authors utilized the Sharpness-Aware Minimization (SAM)~\cite{sam} with the Adam optimizer, which significantly impacted the final scores. SAM is a technique that considers the ``sharpness" of the loss manifold in each gradient descent step, resulting in improved generalization. In SAM, the sharpness is computed from a fixed-size neighborhood of the current location in the loss manifold. However, the fixed-size neighborhood is not always ideal for sharpness computation. Adaptive SAM (ASAM)~\cite{asam} proposed a size-varying neighborhood based on the scale to improve upon it. Taking a cue from this, we propose to utilize ASAM with AdamW optimizer to optimize our proposed framework. Through ablation analysis, we validate the effectiveness of ASAM in boosting the performance of the proposed framework.
 
\section{Experimental Analysis}
This section provides detailed information about the implementation, datasets, followed by the evaluation and ablation study.
\subsection{Implementation Details}
We choose RegNetY200 initialized with pre-trained ImageNet weights as backbone in spatio-temporal feature extractor. 
The model was trained using ASAM~\cite{asam} with AdamW~\cite{adamw} as the base optimizer with $\rho=2$. 
A cosine annealing learning rate scheduler was employed with a warm-up of 3 epochs and an initial learning rate of 1e-5 during the warm-up. For the SoftIC loss, a feature dimension of 128 and a memory bank size of 256 were used, with a loss weight (i.e., $\lambda_{SIC}$) of $0.001$. Mixup~\cite{mixup} augmentation was applied to the training samples with a mixing factor of $\alpha=0.1$. Additionally, random horizontal flips, random Gaussian blur, and random adjustments to brightness, contrast, and saturation-based augmentations were used in the training process. 
The proposed method is implemented in PyTorch and trained on 4 A100 GPUs with a batch size of 2 on each GPU.
At a time, 128 frames were processed, constituting a single sample (clip). During training, clips were uniformly sampled from videos, while during evaluation, a sliding window with half overlap was utilized to extract the samples.

To validate the efficacy of the proposed method, we conducted experiments using event spotting datasets outlined in~\cite{e2espot} such as Tennis~\cite{vid2player}, Figure Skating (FS)~\cite{fs}, Fine Gym and the SoccerNet V2~\cite{SoccerNetv2} action spotting dataset. These datasets are described in the supplementary material.

\subsection{Evaluation Metric Description}
Like previous works, we calculate the average precision (AP) within a specified tolerance range, denoted as $\delta$. The AP is then calculated for each event class to calculate the mean average precision (mAP) with the specified $\delta$ value as an evaluation metric. The $\delta$ is defined based on the respective datasets. For the Tennis and Figure Skating datasets, which have very strict evaluation criteria, we report the results with $\delta=0, 1,$ and $2$. The SoccerNet V2 dataset, which has relaxed evaluation criteria, define two metrics: tight-mAP and loose-mAP, with $\delta=1-4$ seconds and $\delta=5-60$ seconds, respectively. Although the metrics are less strict compared to others, the SoccerNet V2 dataset is more challenging, having a more stringent metric would make the task even harder.

\subsection{Baseline Methods}
For the SoccerNet V2~\cite{SoccerNetv2} dataset, we compared existing SOTA methods proposed in conjunction with the SoccerNet Action Spotting challenge. Some of the results are taken from the 2021 challenge leaderboard, including
Baidu (TF)~\citep{baidu}, 
and SpotFormer~\cite{spotformer}. We have also compared the results with CNN-based SOTA methods like E2E-Spot~\cite{e2espot}, T-DEED~\cite{T-DEED}, UGLF~\cite{uglf} and transformer-base SOTA methods like Spivak~\cite{yahoo-challenge}, ASTRA~\cite{astra}, and COMEDIAN~\cite{comedian}. 
The results were generated using the checkpoints provided by the respective authors whenever possible. 

For the Tennis and Figure Skating dataset, we compared our method with the baseline method E2E-Spot~\cite{e2espot}. We also included the results of some best-performing pre-trained and fine-tuned variants as defined in \cite{e2espot}, namely TSP \cite{tsp} + MS-TCN \cite{ms-tcn}, TSP \cite{tsp} + ASFormer \cite{asformer}, VC-Spot \cite{e2espot}, and 2D-VDP \cite{2d-vdp} + MS-TCN \cite{ms-tcn}. We have also compared the result with recent state-of-the-art T-DEED~\cite{T-DEED} method. Unless otherwise stated, the results are generated mainly from the checkpoint provided by the authors. 

\subsection{Result Analysis on Various Sports Datasets}
This section compares the results obtained by our proposed method with existing SOTA methods on SoccerNet V2, Tennis, Figure Skating \& FineGym datasets. 

\noindent \textbf{SoccerNet V2:}
Table \ref{tab:result-soccernet} presents a comparison between the existing SOTA methods and the proposed approach on the SoccerNet V2 testing dataset. It is evident from the results that the proposed method has achieved notably strong performance in both tight and loose settings. The scores demonstrate significant improvement compared to both RegNet-Y 200MF and 800MF of E2E-Spot, despite utilizing a similar framework. Additionally, there is a substantial improvement in the tight setting as compared to Baidu, Spivak and ASTRA methods. These methods employ a pre-trained frozen larger backbone for feature extraction and transformer-based long-term temporal modules. COMEDIAN model utilized a transformer network to achieve SOTA result in tight setting.
However, the proposed method outperforms these methods without requiring a larger network. Notably, the propose model achieves 2.41\% improvement in tight setting than COMEDIAN model while having significantly less parameters and FLOPs. 

\begin{table}[t!]
\caption{Result Comparison on SoccerNet-V2 datasets in terms of mAP tight and loose settings. Here, the best-performing measures are highlighted with \textbf{bold font} and second best-performing is highlighted with \underline{\textit{underline font}}. * result obtained from the challenge leaderboard of 2021.  ** indicate that the GFLOPs value is calculated from the temporal network only without feature extractors. $\dagger$ indicate result generated from the author provided checkpoint. $\ddagger$ indicate results obtained from official repo.}
\label{tab:result-soccernet}
\centering
\begin{adjustbox}{max width=0.99\linewidth}
\begin{tabular}{lcccc}
\toprule
Model                                            & Params (M)           & GFLOPs & Tight & Loose \\ \midrule
Baidu (TF)~\cite{baidu} & --- & --- & 47.05$^*$ & 73.77$^*$                               \\
SpotFormer~\cite{spotformer} &     ---                 &  ---      & 60.90                                                                    & 76.10                                    \\
E2E-Spot (RegNet-Y 200MF)~\cite{e2espot} & 4.46 & 39.61 & 61.19 & 73.25     \\
E2E-Spot (RegNet-Y 800MF)~\cite{e2espot} & 12.64 & 151.40 & 61.82 & 74.05 \\
Spivak~\cite{yahoo-challenge}& 17.46 & 461.89 &  65.10 & \textit{\underline{78.50}} \\
ASTRA~\cite{astra}& 44.33 & 8.83$^{**}$ & 66.63 & 78.14 \\
T-DEED (RegNet-Y 200MF)~\cite{T-DEED}& 16.36 & 21.96 & 39.43$^{\ddagger}$ & 52.93$^{\ddagger}$ \\
T-DEED (RegNet-Y 800MF)~\cite{T-DEED}& 46.22 & 60.25 & 21.57$^{\ddagger}$ & 30.49$^{\ddagger}$ \\
UGLF~\cite{uglf}& 4.46 & 39.61 & 63.51 & 73.98$^{\ddagger}$ \\
COMEDIAN (ViSwin-T)~\cite{comedian}& 70.12 & 222.76 & \underline{\textit{71.33}}$^{\dagger}$ & 77.07$^{\dagger}$ \\
\rowcolor{lightmintbg} Proposed & 6.46 & 60.25 & \textbf{73.74}                                                           & \textbf{79.11}                          \\
\bottomrule
\end{tabular}
\end{adjustbox}
\end{table}

\begin{table*}[t!]
\caption{Result Comparison on Tennis, FS-Comp, FS-Perf and FinGym datasets in terms of mAP for different delta values. Here, the best-performing measures are highlighted with \textbf{bold font} and second best-performing is highlighted with \underline{\textit{underline font}}. * indicate result generated using provided checkpoints.}
\label{tab:result-tennis-FS}
\centering
\begin{adjustbox}{max width=0.99\linewidth}
\begin{tabular}{lcccccccccccc}
\toprule
\multirow{2}{*}{Model} & \multicolumn{3}{c}{Tennis} & \multicolumn{3}{c}{FS-Comp} & \multicolumn{3}{c}{FS-Perf} & \multicolumn{3}{c}{FineGym} \\ \cmidrule(l){2-13} 
& $\delta$ = 0 & $\delta$ = 1 & $\delta$ = 2 & $\delta$ = 0 & $\delta$ = 1 & $\delta$ = 2 & $\delta$ = 0 & $\delta$ = 1 & $\delta$ = 2 & $\delta$ = 0 & $\delta$ = 1 & $\delta$ = 2 \\ \midrule
TSP~\cite{tsp} + MS-TCN~\cite{ms-tcn}                                     & ---                                                           & 90.90                                   & 95.10                                                         & ---                                                           & 72.40                                   & 87.80                                   & ---                                                           & 76.80                                   & 89.90                                   & --- & 40.50  & 58.50 \\
TSP~\cite{tsp} + ASFormer~\cite{asformer}                                   & ---                                                           & 89.80                                   & 95.50                                                         & ---                                                           & 77.70                                   & 94.10                                   & ---                                                           & 80.20                                   & 94.50                                   & --- & 38.80 & 57.60 \\
2D-VDP~\cite{2d-vdp} + MS-TCN~\cite{ms-tcn}                                  & ---                                                           & ---                                     & ---                                                           & ---                                                           & 83.50                                   & \textbf{96.20}                          & ---                                                           & 85.20                                   & \textit{\underline{96.40}} & --- & --- & --- \\
VC-Spot (RegNet-Y 200MF)~\cite{e2espot} & ---                                                           & 92.40                                   & 96.00                                                         & ---                                                           & 61.80                                   & 75.50                                   & ---                                                           & 56.20                                   & 75.30                                   & --- & 18.70 & 28.60 \\
E2E-Spot (RegNet-Y 200MF)~\cite{e2espot}& \underline{\textit{45.34}}* & \textit{\underline{96.10}} & 97.70                                                         & 22.08*                                                        & 81.00                                   & 93.50                                   & 22.33*                                                        & 85.10                                   & 95.70                                   & 10.01* & 47.90 & 65.20 \\
E2E-Spot (RegNet-Y 800MF)~\cite{e2espot}& 44.64*                                                        & 95.16*                                  & \underline{\textit{97.72}}* & 23.95*                                                        & 83.40                                   & 94.90                                   & 22.17*                                                        & 83.30                                   & 96.00                                   & 10.15* & 50.11* & 67.23* \\
T-DEED (RegNet-Y 200MF)~\cite{T-DEED}& 43.49 & 96.17 & 97.95 & \underline{\textit{28.92}}* & \textit{\underline{85.15}} & 91.70                                   & \underline{\textit{31.87}}* & 86.79                                   & 96.05                                   & \underline{\textit{13.54}}* & \underline{\textit{53.73}}* & 66.43* \\
T-DEED (RegNet-Y 800MF)~\cite{T-DEED}& 44.08 & 96.42 & 97.05 & 27.59*                                                        & 84.77                                   & 92.86                                   & 31.25*                                                        & \textit{\underline{88.17}} & 95.87                                   & 12.42* & \textbf{54.00}* & \underline{\textit{66.70}}* \\
UGLF~\cite{uglf} & ---- & --- & ---- & --- & --- & --- & --- & --- & --- & --- & 50.20 & \textbf{67.80} \\
\rowcolor{lightmintbg} Proposed & \textbf{61.01}                                                & \textbf{96.21}                          & \textbf{97.75}                                                & \textbf{37.54}                                                & \textbf{86.67}                          & \textit{\underline{95.53}} & \textbf{40.36}                                                & \textbf{89.57}                          & \textbf{97.38}                          & \textbf{15.24} & 52.31 & 66.57 \\ \bottomrule
\end{tabular}

\end{adjustbox}
\end{table*}
\begin{figure*}[t]
\subfloat[\scriptsize Tennis dataset \cite{vid2player}]{\centering
    \includegraphics[width=0.49\textwidth]{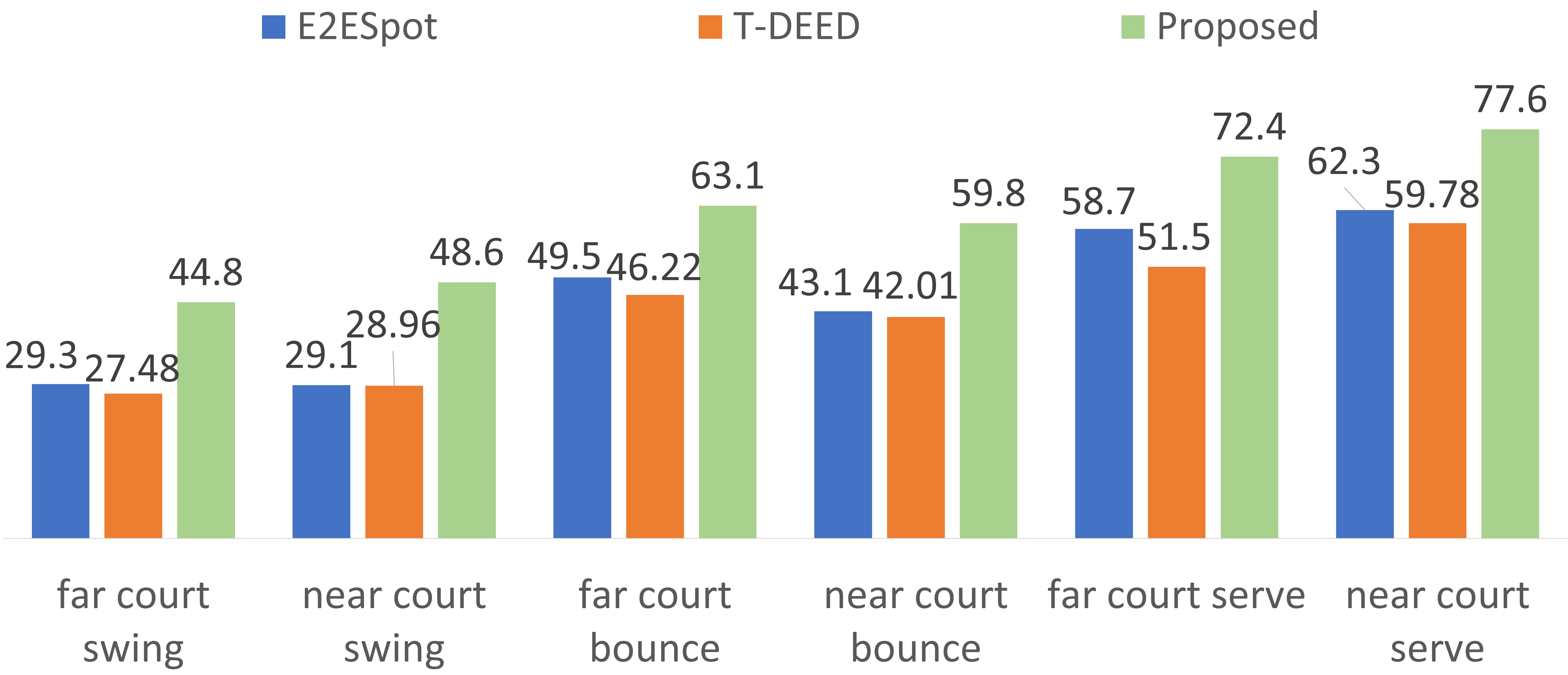}}\hfill
\subfloat[\scriptsize FineGym dataset \cite{finegym}]{\centering
    \includegraphics[width=0.49\textwidth]{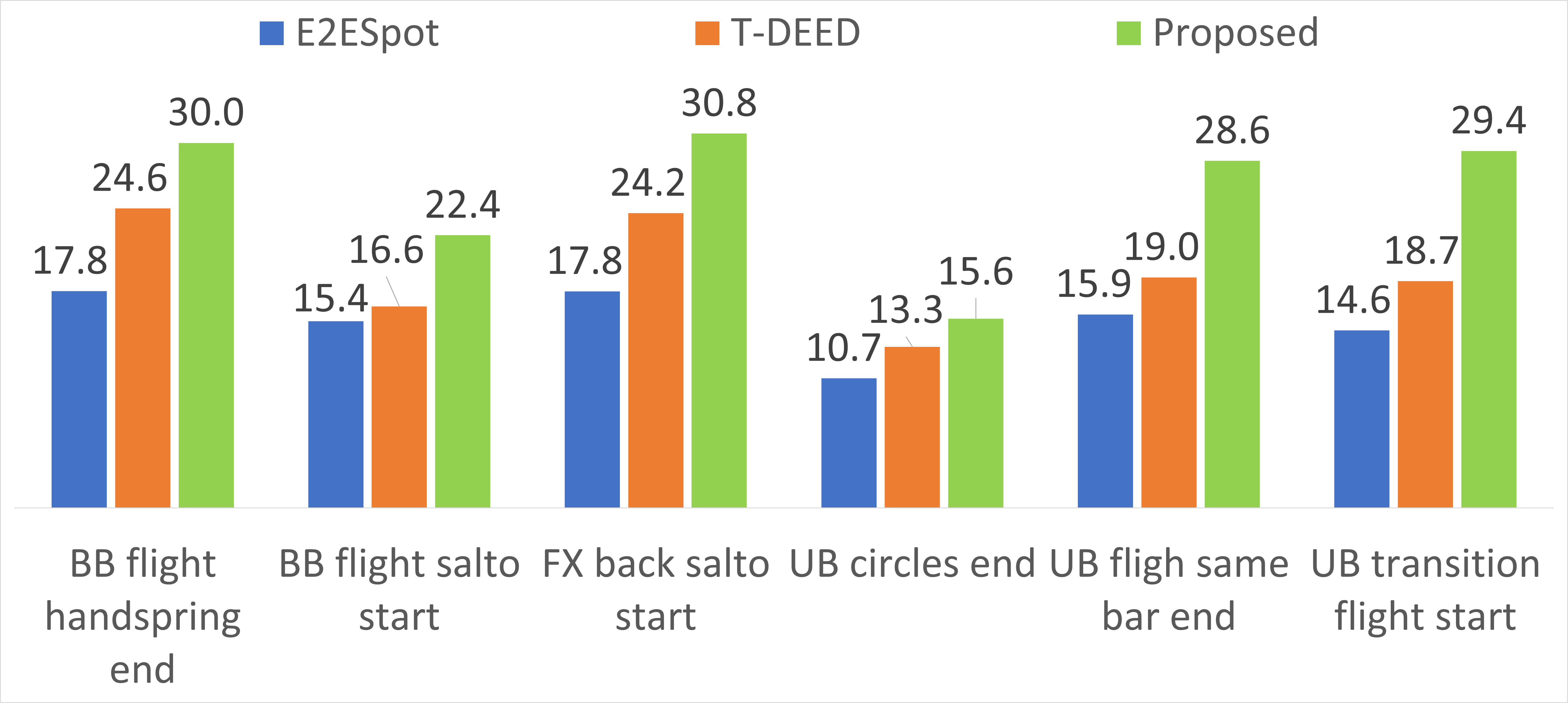}}\\
\subfloat[\scriptsize FS-Comp dataset \cite{fs}]{\centering
    \includegraphics[width=0.45\textwidth]{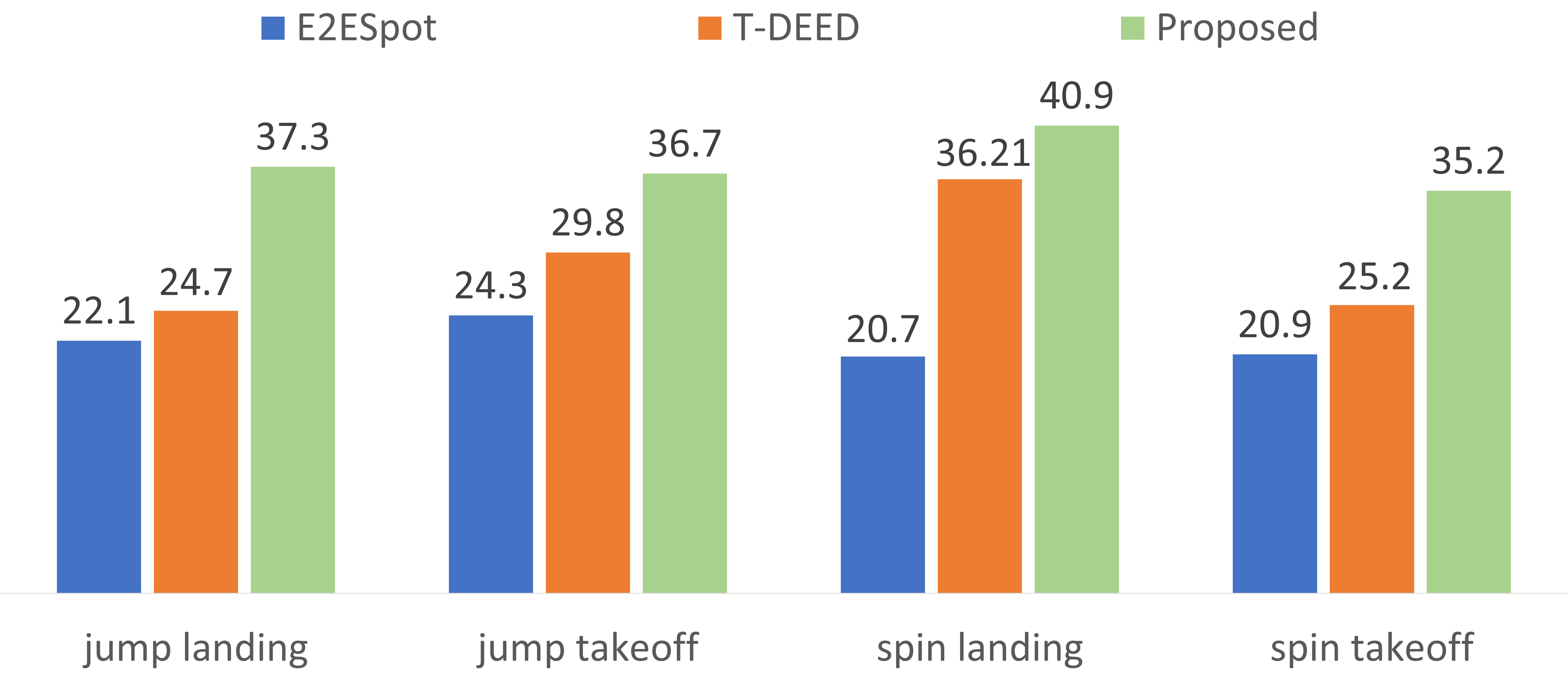}}\hfill
\subfloat[\scriptsize FS-Perf dataset \cite{fs}]{\centering
    \includegraphics[width=0.45\textwidth]{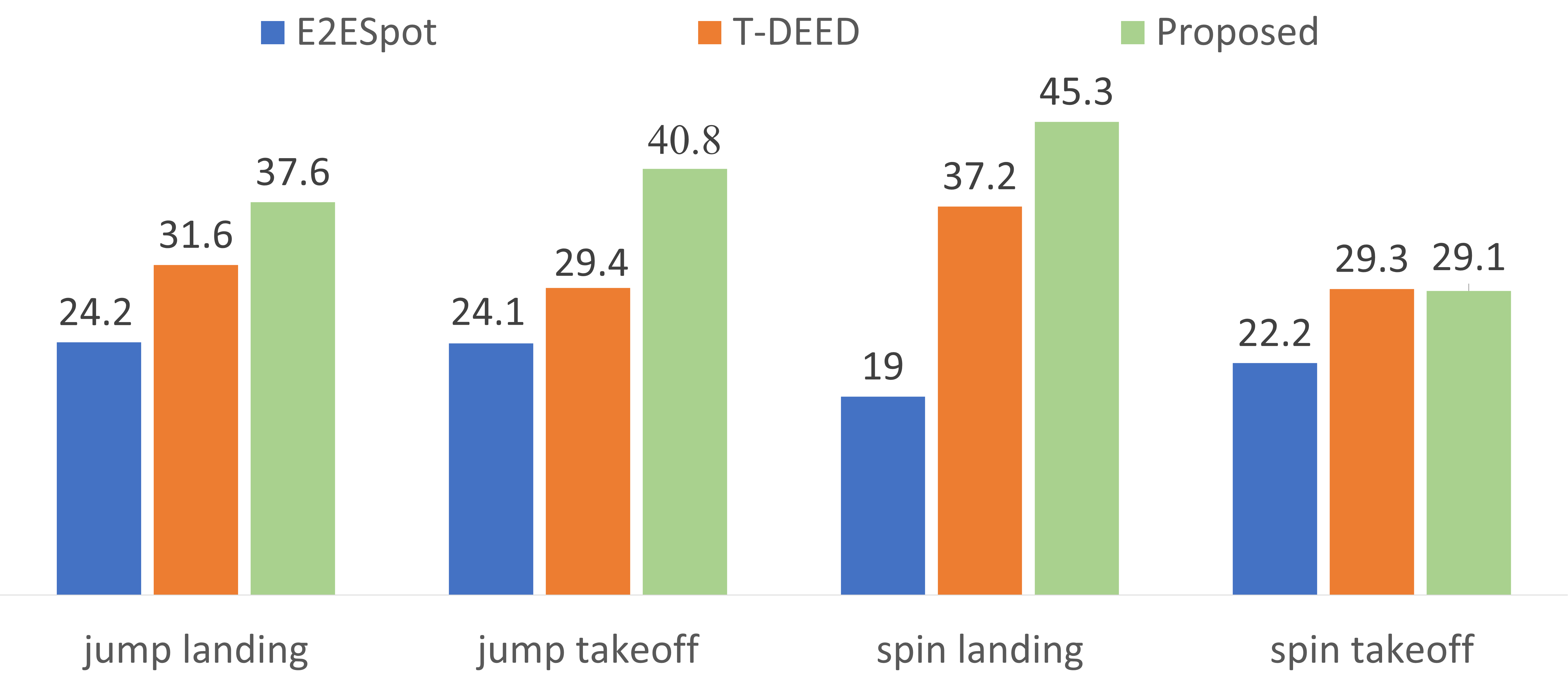}}
\caption{Per-class score comparison on $\delta=0$ setting in terms of \textbf{mAP} for Tennis, FineGym, FS-Comp, and FS-Perf dataset. For FineGym, we report results for only 6 out of 32 classes, while results for all classes in the other datasets are reported.}
    \label{fig:per_class_tennis_fs}
\end{figure*}

\noindent \textbf{Tennis, Figure Skating \& FineGym:}
Table~\ref{tab:result-tennis-FS} shows the results for Tennis, Figure Skating settings, i.e., FS-Comp and FS-Perf, and FineGym with $\delta=0, 1$ and $2$. 
In this case too, the proposed method outperforms existing methods with a noticeable improvement. For instance, in case of $\delta=0$, the proposed method achieves 15.67\%, 8.64\%, 8.49\% and 12.55\% improvements over the best-performing E2E-Spot (RegNet-Y 200MF) and T-DEED (RegNet-Y 200MF) methods in Tennis, FS-Comp \& FS-Perf and FineGym datasets, respectively. Similarly, for $\delta = 1$ case, the proposed method outperforms the previous best-performing T-DEED variants RegNet-Y 200MF and RegNet-Y 800MF by 1.52\% \& 1.4\% improvement on FS-Comp \& FS-Perf datasets, respectively. 
Additionally, we present the per-class score comparison on Tennis \& Figure Skating datasets in Figure~\ref{fig:per_class_tennis_fs}. Here, it can be seen that the performance of the proposed method is not just good for a single class; it has performed consistently well across all the classes with significant margin.

\subsection{Ablation Studies}
\label{sec:ablation}
We conducted ablation studies to validate the design choices made in the proposed framework. To ensure a fair comparison, we analyzed the experimental results on SoccerNet V2 testing dataset. 

\begin{table}[t!]
\centering
\caption{Ablation studies on various aspects of the proposed framework on SoccerNet V2 testing dataset in terms of mAP.}\label{tab:ablation}
\small
\begin{adjustbox}{max width=0.96\linewidth}
\begin{tabular}{@{}lcc@{}}
\toprule
Method & Tight    & Loose    \\
\midrule
\multicolumn{3}{l}{\textbf{Analysis on proposed modules}} \\
\midrule
Baseline: E2E-Spot (RegNet-Y 200MF + GSM) & 61.19  & 73.25        \\
RegNet-Y 200MF + ASTRM & 68.63          &  75.57            \\
RegNet-Y 200MF + ASTRM + SAM  & 71.89  &  77.50            \\
RegNet-Y 200MF + ASTRM + ASAM & 72.65  &  78.58            \\
RegNet-Y 200MF + ASTRM + ASAM + Focal Loss & 70.36  & 76.86      \\
RegNet-Y 200MF + ASTRM + ASAM + IC Loss & 72.05 & 78.30           \\
\rowcolor{lightmintbg} RegNet-Y 200MF + ASTRM + ASAM + SoftICloss & \textbf{73.74}  &  \textbf{79.11}            \\ 
\midrule 
\multicolumn{3}{l}{\textbf{Analysis on proposed ASTRM module}} \\
\midrule
With local spatial only & 69.91 & 76.39 \\
With local temporal only & 70.41 & 76.79 \\
With global temporal only & 71.65 & 77.06 \\
With local spatial + local temporal & 71.06 & 77.10 \\
With local spatial + global temporal & 71.82 & 77.85 \\
With local temporal + global temporal & 71.55  & 76.93 \\
\rowcolor{lightmintbg} Proposed ASTRM module & \textbf{73.74}  &  \textbf{79.11}  \\
\midrule 
\multicolumn{3}{l}{\textbf{Analysis on Different Temporal networks}} \\
\midrule
Baseline with Bi-GRU &  61.19  & 73.25 \\
\rowcolor{lightmintbg} Proposed with Bi-GRU  & \textbf{73.74}   & \textbf{79.11} \\
\quad with Deformable Attention  &  67.43 & 72.15 \\
\quad with Bi-GRU 2 Layers  &  67.85 & 73.03  \\
\quad with Transformer (L1H8)  & 65.89  &  70.76 \\
\quad with Transformer (L2H8)  &  67.60 & 72.35 \\
\quad with Bi-LSTM  &  68.15 & 73.35 \\
\quad with MSTCN  & 68.75  & 73.50  \\
\midrule 
\multicolumn{3}{l}{\textbf{Analysis on Clip Length}} \\
\midrule
Clip length = 64  & 43.82  &  52.09  \\
Clip length = 100 (Used in baseline E2E-Spot) &  71.20 &  76.73  \\
\rowcolor{lightmintbg} Clip length = 128 (proposed) & \textbf{73.74}  &  \textbf{79.11}  \\
Clip length = 192  & 71.74 & 77.02   \\
\bottomrule
\end{tabular}
\end{adjustbox}
\end{table}

\noindent \textbf{Importance of the proposed modules:}
Table~\ref{tab:ablation} shows the contribution of the proposed ASTRM, SoftIC loss and ASAM components to achieve SOTA results.
We observe a significant improvement in scores when replacing GSM (used in the E2E-Spot method) with the proposed ASTRM, highlighting the importance of effective feature extraction. SAM~\cite{sam}, utilized in Spivak method, further enhances the results by a notable margin. Furthermore, we have illustrated how ASAM outperforms the scores achieved by SAM. Although ASAM and SoftIC loss improves the scores by a relatively small amount, each component plays a role in improving the results. 
This emphasizes the contributions of all components to achieve SOTA result.


\noindent \textbf{Components of the proposed ASTRM:}
The proposed ASTRM module is made up using 3 major components: local spatial, local temporal and global temporal. Here, we highlights the importance of each component and the corresponding findings are reported in Table~\ref{tab:ablation}. Notably, the score is at its lowest when utilizing only the spatial aspect of ASTRM; it fails to capture temporal information. Furthermore, the performance of local temporal is superior to using only the spatial aspect. Interestingly, we also observed that global temporal performs better than local temporal alone. This finding supports the notion that observing distant frames may be necessary to classify the current frame accurately. A similar pattern is seen when comparing local spatial + local temporal with local spatial + global temporal. Moreover, the inclusion of local spatial information leads to improved scores. Conversely, using only temporal information results in even lower scores, indicating the necessity of spatial information for understanding ``what is happening" and temporal information for determining ``when it is happening". Overall, our observations highlight the need for spatial and local-global temporal information to identify events precisely and outperform other settings significantly. 

\noindent \textbf{Analysis on Different Temporal modules}
In our proposed approach, we opted for the Bidirectional GRU (Bi-GRU) as the long-range dependency module in the temporal block after conducting various experiments with different potential temporal networks. Table~\ref{tab:ablation} presents the results obtained from the proposed approach using different temporal networks on the SoccerNet V2~\cite{SoccerNetv2} and Tennis~\cite{vid2player} test sets. The results indicate that the Bi-GRU, as a temporal network, performs favorably compared to other temporal networks. Even more advanced temporal networks, such as and Bi-LSTM, Transformer heads, Deformable Attention~\cite{tadtr} and MSTCN~\cite{ms-tcn}, could not surpass the scores achieved by the Bi-GRU temporal network. The difference in scores is visible across all tolerances.  
It is also observed that the proposed method generates significantly better results than the baseline E2E-Spot method, which also uses Bi-GRU as the temporal block. For instance, the proposed method achieves a 12.55\% and 5.86\% improvements over baseline model in the tight and loose settings of SoccerNet V2 dataset, respectively. Similarly, there is an improvement of 15.67\% in $\delta=0$ setting compared to baseline on Tennis dataset. 
Here, this significant improvement is attributed to the proposed ASTRM module and the proposed SoftICLoss function.

\noindent \textbf{Clip Length:}
Due to memory and computation constraints, it's not feasible to process all the frames simultaneously. Therefore, we process a fixed number of consecutive frames at a time, which we refer to as the ``clip length". The value of the clip length affects the scores because it determines how many frames the network considers at once. While we might want to maximize the clip length, we need to find a balance between performance and clip length. The results in Table~\ref{tab:ablation} support this idea. If the clip length is too short, it significantly impacts the results, while making it too long also has a negative effect on the results.

\section{Conclusion \& Future Directions}
This paper proposes an end-to-end network for precise spotting events in sports videos. It effectively tackles the long-range dependency and class imbalance issues and outperforms existing methods. To this extent, we introduce a new module called ASTRM for enhancing spatio-temporal features. Further, to address the class imbalance issue, we propose SoftIC loss, which enforces compact features and improves class separability. We further employed ASAM during training to ensure the model performs well on unseen data.
While our method has advanced the state-of-the-art, especially in the more challenging cases with a relatively simple design, there is further scope for improvement on the two fronts we tried to address. Certain larger networks can improve the performance further but at the cost of computation. Processing the SoccerNet V2 dataset at 2 FPS and 224p is likely to impact the results. It will be interesting to see whether scores improve at higher resolution and FPS. There will be more data but the computation will also increase.

{
    \small
    \bibliographystyle{ieeenat_fullname}
    \bibliography{main}
}


\end{document}

%% file: preamble.tex
\usepackage{lineno}

%% file: sec/0_abstract.tex
\begin{abstract}
Precise Event Spotting (PES) aims to identify events and their class from long, untrimmed videos, particularly in sports. The main objective of PES is to detect the event at the exact moment it occurs. Existing methods mainly rely on features from a large pre-trained network, which may not be ideal for the task. Furthermore, these methods overlook the issue of imbalanced event class distribution present in the data, negatively impacting performance in challenging scenarios. This paper demonstrates that an appropriately designed network, trained end-to-end, can outperform state-of-the-art (SOTA) methods. Particularly, we propose a network with a convolutional spatial-temporal feature extractor enhanced with our proposed Adaptive Spatio-Temporal Refinement Module (ASTRM) and a long-range temporal module. The ASTRM enhances the features with spatio-temporal information. Meanwhile, the long-range temporal module helps extract global context from the data by modeling long-range dependencies. To address the class imbalance issue, we introduce the Soft Instance Contrastive (SoftIC) loss that promotes feature compactness and class separation. Extensive experiments show that the proposed method is efficient and outperforms the SOTA methods, specifically in more challenging settings.


\end{abstract}

%% file: sec/1_intro.tex
\section{Introduction}
\label{sec:intro} 
Event spotting methods locate the occurrence of events in long, untrimmed videos. An event is typically defined by a starting and ending time, but this may not always be feasible, particularly in the context of sports, where an event is a specific occurrence that occurs at a particular point in time, as dictated by the game's rules. For instance, a ``Goal" in soccer is scored when the ball crosses the goal line, while a ``Boundary" in cricket is made when the ball touches the boundary line. This is termed as Precise Event spotting (PES) and PES methods \cite{e2espot,SoccerNetv2} aim to predict an event at the exact frame it occurs (e.g., goal, boundary). This is highly relevant in generating summaries of the game, finding specific events, and editing long matches. However, it is a challenging task since it requires a high-level understanding of the game and its key events. 
\begin{figure}
    \centering
    \includegraphics[width=0.99\linewidth]{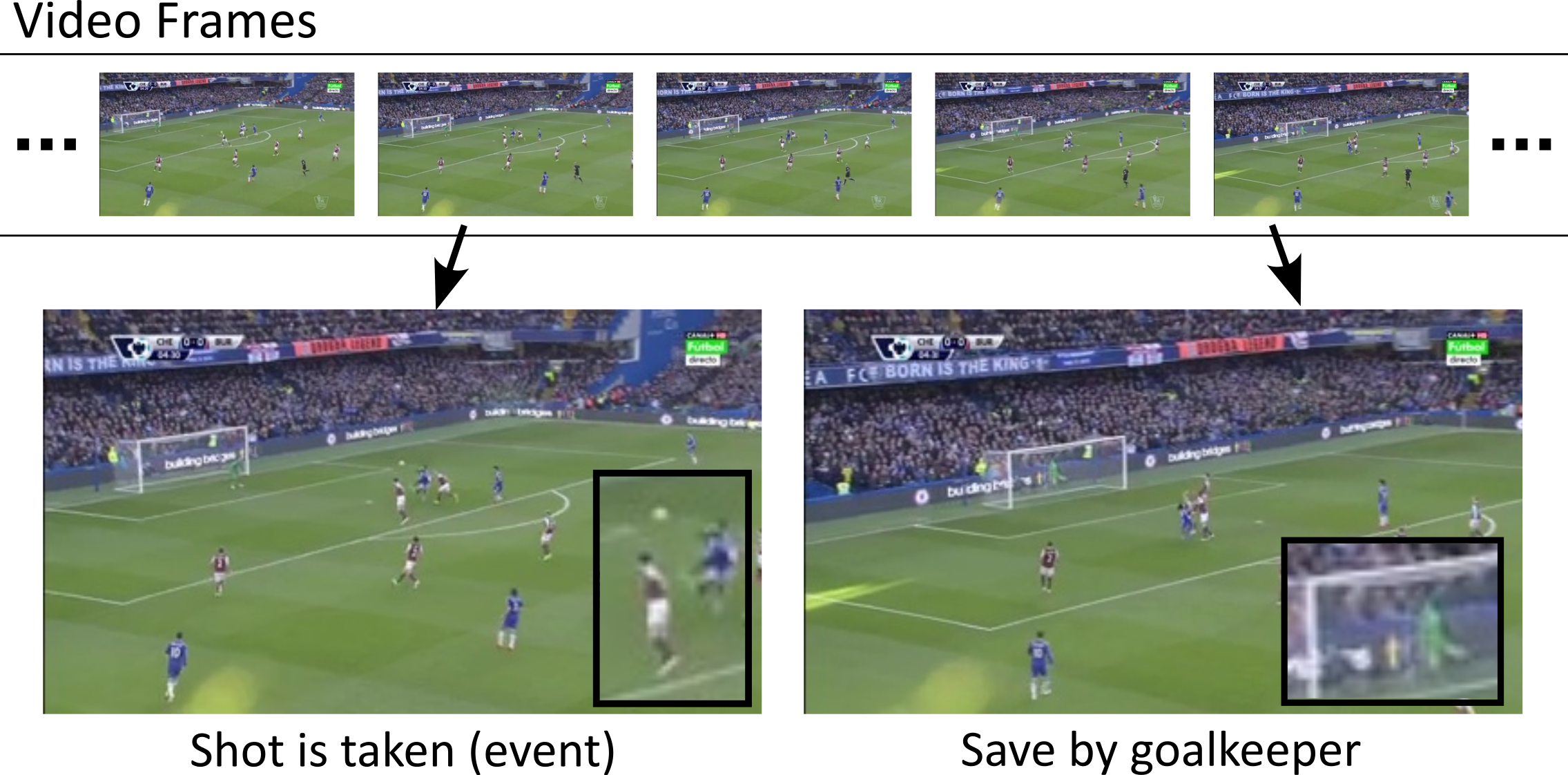}
    \caption{\textbf{Temporal dependency of ``Shots on target'' event}: The event is marked when the shot is taken, but whether the shot is at the target can only be known by looking at future frames. Here the shot is \emph{on target} because the ball is saved by the goal keeper.}
    \label{fig:long_range_dependency}
\end{figure}
\begin{figure}
    \centering
    \includegraphics[width=\linewidth]{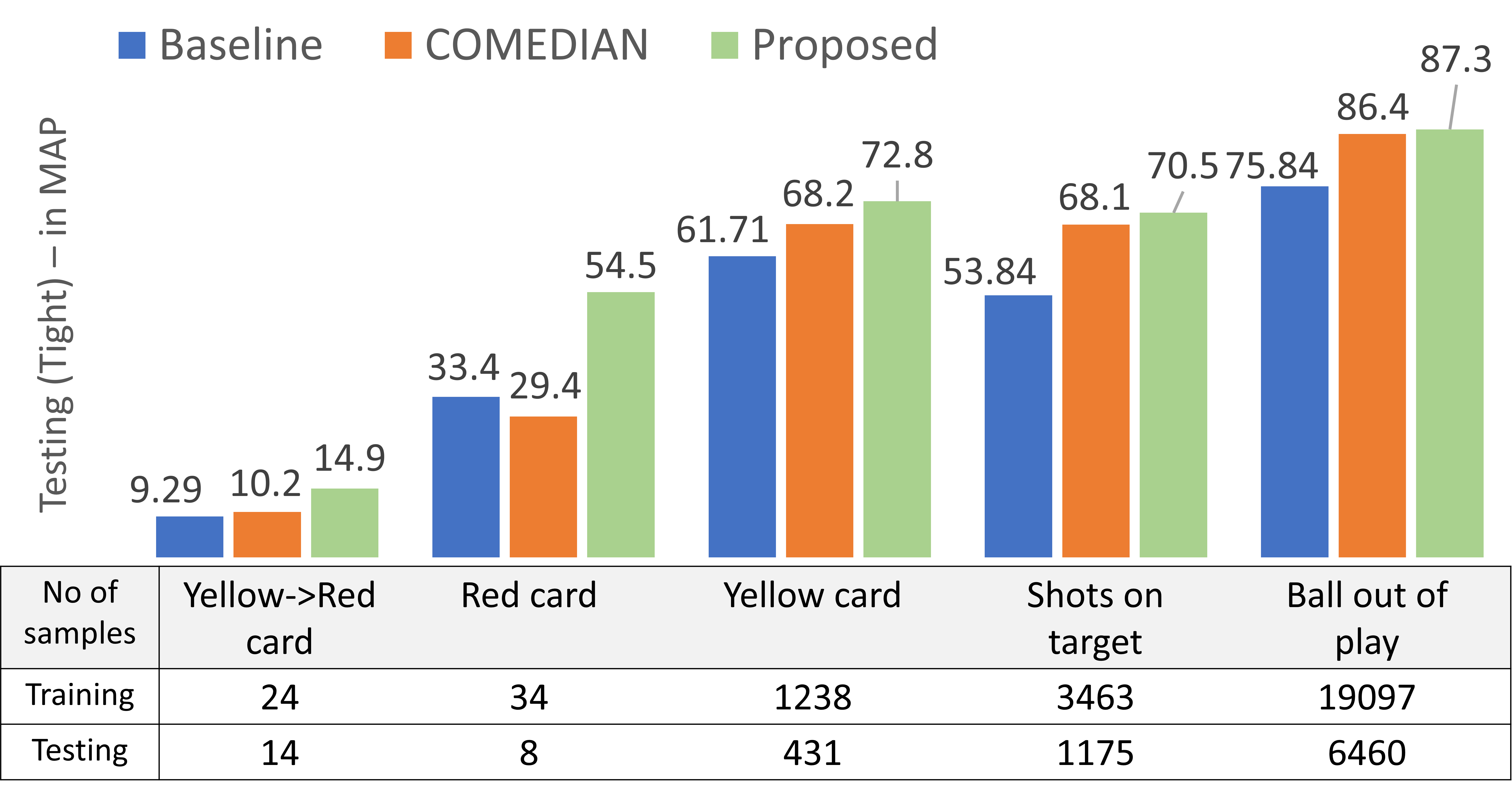}
    \caption{\textbf{Per class score analysis on a  few classes of SoccerNet V2 dataset in tight setting.} The proposed method outperforms state-of-the-art (SOTA) methods especially for classes with less number of samples.}
    \label{fig:sample_score_graph}
\end{figure}

Existing PES methods mainly face two challenges. First, the long-range temporal dependency that arises due to nature of events. 
To predict the event frames observing just the nearby frames does not suffice, several other potentially distant frames need to be observed to correctly predict the event.
For example, ``Shots on target'' and ``Shots off target'' events of Soccer game are marked when the shot is taken, but the actual class is decided based on whether the ball went towards the goal or away from it~(see Figure~\ref{fig:long_range_dependency}). This can be only detected if long-range dependencies are captured correctly. Second, there is an inherent imbalance in the number of samples among the classes~(see Figure~\ref{fig:sample_score_graph}). Some events naturally occur more frequently than others. For example, in a soccer game, a ``Ball out of play'' is likely to occur more than a ``Yellow card''; a ``Red-card'' and ``Yellow\textrightarrow Red card'' are expected to occur even less often. 
Therefore, the same trend can be observed in datasets created using broadcast videos. 
Due to these issues, existing event spotting methods \cite{e2espot, comedian} perform poorly in underrepresented classes such as ``Red card" \& ``Yellow$\rightarrow$Red card." This can be visualized in Figure~\ref{fig:sample_score_graph}. Hence, handling these issues and accurately performing PES is an active area of research.

To address the long-range temporal dependency issue, recent SOTA methods~\cite{yahoo-icip,astra} prefer to use the feature extracted using the pre-trained network of \cite{baidu}. These methods also use transformer-based temporal modules 
due to their better sequence modeling capability with a global context. However, as stated in \cite{asformer, tadtr}, transformers also encounter issues with long videos. Self-attention struggles to learn appropriate weights for long videos due to the increase in the size of the weight matrix with the number of frames being processed. 
On the other hand, E2E-Spot \cite{e2espot} proposed an end-to-end trainable feature extractor and a GRU-based module for temporal modeling. Recently proposed T-DEED~\cite{T-DEED} improves upon the performance of E2E-Spot via temporal-discriminability enhancing encoder decoder (T-DEED) module. In our proposed method, we handle this issue by improving the extracted features with our proposed Adaptive Spatio-Temporal Refinement Module (ASTRM) and GRU-based temporal module. 


The class imbalance issue has not received the sufficient attention in the literature of PES, resulting in lower performance on real-world datasets. The recent method ASTRA~\cite{astra} is one of the few attempts to address this problem through a modified "mixup" augmentation. 
While there are standard machine learning techniques available, such as focal loss~\cite{focal_loss}, these methods are not suitable for every task, as demonstrated in this paper. In response to this challenge, we introduce a novel loss function called Soft Instance Contrastive (SoftIC) loss, which aims to alleviate class imbalance by promoting compact feature representations and enhancing class separability.



Finally, this paper primarily aims to address these issues and improve the accuracy of PES. Here, we propose an end-to-end trainable network with a CNN-based spatio-temporal feature extractor and a GRU-based long-term temporal module. The spatio-temporal feature extractor includes our proposed Adaptive Spatio-Temporal Refinement Module (ASTRM) to augment the features with spatial and temporal information. We chose to train the model end-to-end because it has been shown to perform better that a pre-trained frozen feature extractor~\cite{rmsnet,e2e_tad}. We have also employed the Adaptive Sharpness-Aware Minimization (ASAM)~\cite{asam} to enable the trained model to perform well on unseen data. Through experiments, we demonstrate that the proposed framework obtains superior performance compared to SOTA methods, especially in very tight settings where long-range dependency and class imbalance issues affect a lot (see Figure~\ref{fig:sample_score_graph}). 
Our key contributions can be summarized as follows: 
\begin{itemize}
    \item We propose an end-to-end framework using CNN-based spatio-temporal feature extractor and GRU-based temporal network to achieve precise event spotting while address long-range dependency and class imbalance issues.
    \item We introduce ASTRM, which enhances the features with spatial and temporal information.
    \item We also introduce SoftIC loss to enforce compact features and improve class separability, addressing the class imbalance issue.
\end{itemize}